\documentclass{article}

\usepackage{arxiv}
\usepackage{hyperref}

\usepackage{graphics} 
\usepackage{epsfig} 
\usepackage{mathptmx} 
\usepackage{times} 
\usepackage{amsmath} 
\usepackage{amssymb}  

\usepackage{multirow}
\usepackage{graphicx}
\usepackage{array}

\usepackage{accents}
\usepackage{algorithm}
\usepackage{algpseudocode}
\usepackage{nicematrix}
\usepackage[caption=false]{subfig}
\usepackage{booktabs}
\usepackage{caption}

\newcommand\Tstrut{\rule{0pt}{2.6ex}}
\newcommand\Bstrut{\rule[-1.5ex]{0pt}{0pt}}

\usepackage{xcolor}

\usepackage{amssymb}

\usepackage{amsmath}
\usepackage{mathdots}

\usepackage{mathtools}

\usepackage{tensor}

\usepackage{urwchancal}
\DeclareFontFamily{OT1}{pzc}{}
\DeclareFontShape{OT1}{pzc}{m}{it}{<-> s * [1.10] pzcmi7t}{}
\DeclareMathAlphabet{\mathpzc}{OT1}{pzc}{m}{it}

\usepackage{graphicx}
\usepackage{ifpdf}
\ifpdf
\usepackage{epstopdf}
\PrependGraphicsExtensions{.svg}
\fi

\newtheorem{theorem}{Theorem}[section]
\newtheorem{lemma}[theorem]{Lemma}

\newtheorem{remark}[theorem]{Remark}

\providecommand{\R}{\mathbb{R}}

\providecommand{\SO}{\mathbf{SO}}

\providecommand{\SE}{\mathbf{SE}}
\providecommand{\SOT}{\mathbf{SOT}}

\providecommand{\MR}{\mathbf{MR}}
\providecommand{\SLAM}{\mathbf{SLAM}}

\providecommand{\grpG}{\mathbf{G}}

\providecommand{\gothg}{\mathfrak{g}}

\providecommand{\so}{\mathfrak{so}}

\providecommand{\slam}{\mathfrak{slam}}

\providecommand{\Sph}{\mathrm{S}}

\providecommand{\calSE}{\mathcal{SE}}

\providecommand{\calM}{\mathcal{M}}
\providecommand{\calN}{\mathcal{N}}

\providecommand{\calU}{\mathcal{U}}

\providecommand{\torSE}{\mathcal{SE}}
\providecommand{\torG}{\mathcal{G}}

\providecommand{\cset}[2]{\left\{ {#1} \mid {#2} \right\}}

\providecommand{\tT}{\mathrm{T}} 
\providecommand{\id}{\mathrm{id}} 

\providecommand{\td}{\mathrm{d}}
\providecommand{\tD}{\mathrm{D}}

\providecommand{\ddt}{\frac{\td}{\td t}}

\providecommand{\mr}[1]{\mathring{#1}} 

\usepackage{accents}
\usepackage{mathtools}
\makeatletter
\providecommand{\scirc}{%
    \hbox{\fontfamily{\rmdefault}\fontsize{0.4\dimexpr(\f@size pt)}{0}\selectfont{\raisebox{-0.52ex}[0ex][-0.52ex]{$\circ$}}}}

\makeatother

\makeatletter
\providecommand{\ucirc}{%
    \hbox{\fontfamily{\rmdefault}\fontsize{0.4\dimexpr(\f@size pt)}{0}\selectfont{\raisebox{0.0ex}[0ex][-0.52ex]{$\circ$}}}}

\makeatother

\mathchardef\mhyphen="2D

\providecommand{\etal}{\textit{et al.~}}

\newcommand{\publicationdetails}{\
    Ge, Y., \etal(2025),``Equivariant Filter Design for Range-only SLAM", \textit{Accepted for presentation at IEEE International Conference on Robotics and Automation.}
\\
\copyrightNoticeIEEE{2025} 
}
\newcommand{\publicationversion}
{Author accepted version}

\providecommand{\logG}{\log_\grpG}

\providecommand{\expG}{\exp_\grpG}
\providecommand{\xizero}{\mr{\xi}}
\providecommand{\RSLAM}{\SLAM^R_n(3)}
\providecommand{\rslam}{\slam^R_n(3)}

\title{\LARGE \bf
Equivariant Filter Design for Range-only SLAM
}

\author{
    \href{https://orcid.org/0000-0001-7969-7039}{\includegraphics[scale=0.06]{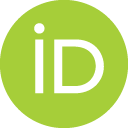}\hspace{1mm}
Yixiao Ge}
\\
    Systems Theory and Robotics Group \\
    School of Engineering \\
	Australian National University \\
    ACT, Australia \\
    \texttt{Yixiao.Ge@anu.edu.au} \\
    \And   {\includegraphics[scale=0.06]{orcid.png}\hspace{1mm}
Arthur Pearce}
\\
    Systems Theory and Robotics Group \\
    School of Engineering \\
    Australian National University \\
    ACT, Australia \\
    \texttt{Arthur.Pearce@anu.edu.au}\\
\And    \href{https://orcid.org/0000-0003-4391-7014}{\includegraphics[scale=0.06]{orcid.png}\hspace{1mm}
Pieter van Goor}
\\
Robotics and Mechatronics (RaM) Group \\
EEMCS Faculty \\
University of Twente \\
Enschede, The Netherlands \\
    \texttt{p.c.h.vangoor@utwente.nl} \\
	\And	\href{https://orcid.org/0000-0002-7803-2868}{\includegraphics[scale=0.06]{orcid.png}\hspace{1mm}
    Robert Mahony}
\\
    Systems Theory and Robotics Group \\
    School of Engineering \\
	Australian National University \\
    ACT, Australia \\
	\texttt{Robert.Mahony@anu.edu.au} \\
}

\begin{document}
\maketitle

\begin{abstract}

Range-only Simultaneous Localisation and Mapping (RO-SLAM) is of interest due to its practical applications in ultra-wideband (UWB) and Bluetooth Low Energy (BLE) localisation in terrestrial and aerial applications and acoustic beacon localisation in submarine applications.
In this work, we consider a mobile robot equipped with an inertial measurement unit (IMU) and a range sensor that measures distances to a collection of fixed landmarks.
We derive an equivariant filter (EqF) for the RO-SLAM problem based on a symmetry Lie group that is compatible with the range measurements.
The proposed filter does not require bootstrapping or initialisation of landmark positions, and demonstrates robustness to the no-prior situation.
The filter is demonstrated on a real-world dataset, and it is shown to significantly outperform a state-of-the-art EKF alternative in terms of both accuracy and robustness.
\end{abstract}

\section{INTRODUCTION}

    Simultaneous Localisation and Mapping (SLAM) refers to the problem of estimating the pose of a mobile robot and the structure of its environment concurrently.
    This problem has been an active reseach topic in robotics for the last 40 years \cite{bailey2006simultaneous}.
    Range-only SLAM (RO-SLAM), also known as range-aided SLAM, refers to the particular case where the only exteroceptive measurements are provided by range-only sensors.
    Its application domains include underwater \cite{paull2013auv}, aerial \cite{guo2017ultra} and ground \cite{funabiki2020range} robotics.

    The RO-SLAM problem is challenging because of the unique nonlinearity introduced by the range measurements, as well as the significant ambiguity \cite{blanco2008efficient}.
    In \cite{kantor2002preliminary}, the authors present preliminary results using an Extended Kalman filter (EKF) for range-only localisation.
    However, because of the special nonlinearity of the range measurements, the EKF implementation requires careful initialisation to ensure the linearisation error does not lead to filter divergence.
    In order to address the case where no prior is available, the authors of \cite{leonard2002mapping} propose a delayed-state EKF which runs a batch process over multiple measurements to identify the optimal linearisation point.
    A similar approach is presented in \cite{olson2006robust} which identifies the linearisation point with the highest likelihood with large data sets.
    In \cite{blanco2008efficient}, the authors propose a fully probabilistic Rao-Blackwellised particle filter for range-only SLAM, which removed the need for delayed initialization or ad-hoc batch initializations.
    In \cite{djugash2009robust}, Djugash and Singh propose the relative-over parametrized EKF (ROP-EKF) which uses polar coordinates to model the range measurements, instead of the conventional Cartesian coordinates.
    Such modelling choices significantly improve the linearisation of the EKF and shows promising results in the no-prior situation.
    However, the associated system kinematics become highly nonlinear and once again the EKF formulation requires careful initialisation to ensure the linearisation error does not lead to filter divergence.
    In parallel to filtering developments, another strand of research treats RO-SLAM as an optimisation problem.
    One seminal work in this direction by Newman and Leonard \cite{newman2003pure} approaches the problem by formulating it as a batch least-squares estimation.
    More recently, Boroson \etal \cite{boroson2020inter} propose a multi-robot SLAM method using range measurements based on pose graph optimisation.
    In \cite{papalia2024certifiably}, the authors present a global optimisation method that provides a certifiable optimal SLAM solution.

A more recent stream of research has investigated symmetry and Lie group structure in other variants of the SLAM problem to develop high-performance filters with improved robustness.
In \cite{barrau2015ekf}, the authors apply the invariant EKF (IEKF) framework \cite{Barrau_2017} to improve the propagation linearisation and the filter consistency.
In \cite{mahony2017geometric}, the authors introduce a quotient manifold structure, termed the SLAM-manifold, which is further exploited in \cite{van2019geometric} to develop a geometric observer for visual SLAM problem.
Van Goor \etal \cite{van2022equivariant} propose the equivariant filter (EqF), a general filter design methodology for systems on Lie groups and homogeneous spaces, which has been applied to visual-inertial SLAM problem \cite{van2023eqvio} and GNSS-aided navigation problem \cite{fornasier2023equivariant} with state-of-the-art performance.
These recent works incorporate a polar-coordinate representation of the visual bearing measurements intrinsically in the state representation.
To the authors' knowledge, these advances in SLAM geometry have yet to be applied to the range-only SLAM problem.

In this paper, we present a novel filter design for RO-SLAM, based on the equivariant filter framework.
Similar to the previous work in IEKF and EqF solutions \cite{van2023eqvio,fornasier2023equivariant,barrau2015ekf} to inertial navigation problems, this work exploits the $\SE_2(3)$ structure for the navigation states of the robot.
Extending existing results, we show that the symmetry Lie group $\RSLAM$ is compatible with range measurements.
The normal coordinates based on the $\SOT(3)$ symmetry \cite{van2023eqvio} provide a natural representation of the nonlinear distributions of the landmark positions in the RO-SLAM problem.
The symmetry structure of the landmark measurement also enables the EqF to use the higher order equivariant output approximation.
Through posing the uncertainty distribution directly on the normal coordinates of $\SOT(3)$, the proposed EqF is robust to the no-prior situation and naturally compatible with landmarks at different ranges.
Finally, we demonstrate the performance of the proposed filter on a real-world dataset for the full SLAM task, and provide a comparison with a standard EKF.

\section{PRELIMINARY}
For a comprehensive introduction to smooth manifolds, Lie groups and their geometric structures, the authors recommend \cite{lee2012smooth}.

Let $\calM$ be a smooth manifold, $\mathrm{T}_\xi\calM$ denote the tangent space at a point $\xi\in\calM$.
Given a differentiable function between smooth manifolds $h:\calM\rightarrow\calN$, its derivative at $\xizero$ is written as
\begin{align*}
    \tD_\xi\big|_{\xizero} h(\xi): \tT_{\xizero}\calM\rightarrow \tT_{h(\xizero)}\calN.
\end{align*}

Let $\grpG$ be a general Lie group with dimension $n$.
Then $\torG$ denotes the $\grpG$-torsor, which is the manifold underlying $\grpG$ stripped of the group structure.
The identity element is denoted $\id \in \grpG$.
The Lie algebra of $\grpG$ is denoted $\gothg$ and is isomorphic to the vector space $\R^n$ with the same dimension.
The exponential and logarithm are written as $\expG:\gothg\to\grpG$ and $\logG:\grpG'\to\gothg$, respectively, where $\grpG'\subset\grpG$ is the subset of $\grpG$ over which the exponential map is invertible.

A (right) group action of a Lie group $\grpG$ on a manifold $\calM$ is a smooth map $\phi:\grpG\times\calM\rightarrow\calM$ that satisfies
\begin{align*}
    \phi(X,\phi(Y,\xi))=\phi(YX,\xi) && \textrm{and} && \phi(\id,\xi)=\xi
\end{align*}
for all $X,Y \in\grpG$ and $\xi\in \calM$.
It induces the partial maps $\phi_X:\calM\rightarrow\calM$ and $\phi_\xi:\grpG\rightarrow\calM$ which are defined by $\phi_X(\xi):=\phi(X,\xi) =: \phi_\xi(X)$.

\subsection{Relevant Lie groups}
The special orthogonal group $\SO(3)$ is the group of all the 3D rotation matrices,
\begin{align*}
    \SO(3) := \cset{R \in \R^{3\times 3}}{R^\top R = I_3 , \; \det(R) = 1}.
\end{align*}
The Lie algebra of $\SO(3)$ is given by
\begin{align*}
    \so(3) &:= \cset{\omega^\times \in \R^{3\times 3}}{\omega \in \R^3}, \quad \text{where}\\
    \omega^{\times}&:=\begin{pmatrix}
        0 & -\omega_3 & \omega_2 \\
        \omega_3 & 0 & -\omega_1 \\
        -\omega_2 & \omega_1 & 0
    \end{pmatrix}
    \quad \forall \omega \in \R^3.
\end{align*}
The Lie group $\SO(3)$ acts on directions by transpose-left multiplication, $\eta \mapsto R^\top \eta$ for $R \in \SO(3)$ and $\eta \in \Sph^2 \subset \R^3$ on the unit sphere.

The extended pose group $\SE_2(3)$ \cite{barrau2015ekf} has elements $T = (A_T,a_T,b_T)\in \SO(3)\ltimes(\R^3\times\R^3)$ with group multiplication $(A_1, a_1, b_1)\cdot(A_2, a_2, b_2) = (A_1 A_2, a_1+A_1 a_2, b_1+A_1 b_1)$.
The matrix representation of $\SE_2(3)$ is given by
\[
    T = \begin{pmatrix}
        A_T & a_T & b_T\\
    \mathbf{0}_{1\times 3} & 1 & 0\\
    \mathbf{0}_{1\times 3} & 0 & 1
    \end{pmatrix}\in\R^{5\times 5}.
\]

The scaled orthogonal transformations $\SOT(3)$ are scaled rotations; that is, transformations that rotate Euclidean space and then scale the whole space homogeneously \cite{van2019geometric}.
Elements of this group are written $Q = c_Q R_Q \in\SOT(3)$, where $R_Q \in \SO(3)$ is a rotation and $c_Q \in \MR(1) \simeq \R_+$ is a scaling factor.
The right action of $\SOT(3)$ on $\R^3$ is
$p \mapsto Q^{-1} q = c_Q^{-1} R_Q^\top p$.

\section{PROBLEM DESCRIPTION}
Consider a mobile robot equipped with an on-board IMU and a range sensor that measures the distances to some beacons.
Choose an arbitrary inertial reference frame $\{0\}$, let $\{B\}$ denote the robot's body-fixed frame.
For simplicity, we assume that both the IMU and range sensor are rigidly attached at the origin of $\{B\}$ and share the same orientation.
The state space of the RO-SLAM problem is identified as $$\calM = \torSE_2(3)\times (\R^3)^n,$$ where $\torSE_2(3)$ is the $\SE_2(3)$-torsor with elements representing the extended pose.
Each element on the state space can be written as $(P,p_1,\cdots,p_n)\in\calM$, where $P = (R, v, x)\in\torSE_2(3)$ represents the navigation states including the orientation $R\in \SO(3)$, the linear velocity $v \in \R^3$ and the position $x \in \R^3$ of the robot with respect to the inertial frame $\{0\}$.
The position of landmark $i$ in the inertial frame $\{0\}$ is denoted by $p_i\in\R^3$.
We will use the notation $(P,p_i) = (R, v, x, p_i) = (R, v, x, p_1, \cdots, p_n)$ as a shorthand throughout the paper.

The gyroscope and accelerometer measurements are written as $\omega\in\R^3$ and $a\in\R^3$, representing the angular velocity and acceleration in $\{B\}$ respectively.
Let $g\in\R$ denote the magnitude of the acceleration due to gravity.
Then the system dynamics are
\begin{align}\label{eq:system_dynamics}
    \ddt(R, v, x, p_i) = f_{(\omega, a)}(R, v, x, p_i),
\end{align}
\begin{align*}
    \dot{R} = R\omega^\times,\qquad
    \dot{v} = R a + g\mathbf{e}_3, \qquad
    \dot{x} = v,\qquad
    \dot{p_i} = 0.
\end{align*}

\begin{remark}
Note that, for simplicity of presentation in this preliminary work, we do not model biases in the IMU measurements.
The proposed symmetry in Section \ref{sec:filter_design} can be easily extended to allow the modelling of the bias states in the filter, see \cite{fornasier2023equivariant} for a comprehensive analysis of symmetry for inertial navigation systems.
\end{remark}

The range sensor measures the distances from the sensor located at $\{ B \}$ to the landmarks $p_1,...,p_n$.
The output function $h:\calM\to\calN$ is thus defined as
\begin{align}\label{eq:measurement_function_global}
    h(R, v, x, p_i):=\left(\lVert p_1 - x\rVert, \cdots,\lVert p_n - x\rVert\right),
\end{align}
with the output space $\calN:=\R_+\times\cdots\times\R_+$.
We assume that the vehicle is never co-located with a beacon.

\subsection{Ego-state reformulation}
For convenience of derivation going forward, we reformulate the dynamics of the landmarks into the body-fixed frame $\{B\}$.
Define the landmark position in $\{B\}$ to be $q_i = R^\top(p_i-x)\in\R^3$ for each $i=1,...,n$.
While the landmark position remains static in $\{0\}$, in the body-fixed frame landmark evolves according to
\begin{align}\label{eq:q_dyanmics}
    \dot{q_i}   &= \ddt\left(R^\top(p_i-x)\right) = \dot{R}^\top(p_i-x)+R^\top(0-v)\nonumber\\
                &= -\omega^\times R^\top (p_i-x) - R^\top v \nonumber\\
                &= -\omega^\times q_i - R^\top v.
\end{align}
Instead of the system state $(R, v, x, p_i)\in\calM$, we may now consider the system state $(R, v, x, q_i)$, where the dynamics are given by \eqref{eq:system_dynamics} and \eqref{eq:q_dyanmics}, and the output can be written
\begin{align}\label{eq:measurement_function_bff}
    h(R, v, x, q_i):=\left(\lVert q_1\rVert, \cdots,\lVert q_n\rVert\right).
\end{align}
For the remainder of the paper, we will consider the system state representation $(R, v, x, q_i)\in\calM$.

\section{EQUIVARIANT FILTER DESIGN}\label{sec:filter_design}

In this section, we develop the proposed EqF design for RO-SLAM according to the procedure of \cite{van2022equivariant}.
We specify a group action of a group we term $\RSLAM$ on the RO-SLAM state space, and derive a lift of the system to the group.
Moreover, we show that the output is equivariant with respect to this group action, enabling us to use the equivariant output approximation $C^\star$ \cite{van2022equivariant}, which eliminates second order terms from the output approximation.
The EqF we propose here also includes the reset step described in \cite{ge2022equivariant}.

\subsection{Symmetry of RO-SLAM}\label{sec:symmetry}
Let $\RSLAM = \SE_2(3)\times\SOT(3)^n$ denote the \emph{Range-SLAM Group}.
Note that the construction of the Lie group $\RSLAM$ is identical to the \emph{VI-SLAM Group} proposed in \cite{van2023eqvio}, but in a different context.
Elements of $\RSLAM$ are written as $X = (T, Q_1, \ldots, Q_n) \in\RSLAM$, and the group product, identity and inverse are given by
\begin{align*}
    (T^1, Q_i^1)\cdot(T^2, Q_i^2) &= (T^1 T^2, Q_i^1 Q_i^2),\\
    \id = (I_5, (I_4)_i),\qquad & (T, Q_i)^{-1} = (T^{-1}, Q_i^{-1}),
\end{align*}
where we once again rely on the shorthand $(T,Q_i) = (T, Q_1, \ldots, Q_n)$ with $i = 1, \ldots, n$.

The symmetry actions and system lift required for the EqF design are given in the following lemmas.
The proofs of Lemma~\ref{lemma:output_action} and \ref{lemma:lift} are provided in Appendix~\ref{sec:proofs}.

\begin{lemma}[{\cite[Lemma 4.2]{van2019geometric}}]
    \label{lemma:state_action}
    The group action $\phi:\RSLAM\times\calM\to\calM$ defined by
    \begin{align}\label{eq:state_action}
        \phi((T, Q_i), (P,  &q_i)) :=(P T, c_{Q_i}^{-1}R_{Q_i}^\top q_i),
    \end{align}
    is a transitive right group action.
\end{lemma}

\begin{lemma}\label{lemma:output_action}
The map $\rho:\RSLAM\times\calN\to\calN$ defined by
\begin{align}\label{eq:output_action}
\rho((T, Q_i),(\lVert q_i \rVert)):= (c_{Q_i}^{-1}\lVert q_i \rVert),
\end{align}
is a transitive right group action.
Moreover, the output function~\eqref{eq:measurement_function_bff} is equivariant with respect to \eqref{eq:state_action} and \eqref{eq:output_action}; i.e.
\begin{align}\label{eq:output_equivariance_condition}
    h(\phi((T, Q_i), (P,  q_i))) = \rho((T, Q_i), h(P,  &q_i)),
\end{align}
for any $(T, Q_i)\in\RSLAM$ and $(P,  q_i)\in\calM$.
\end{lemma}

\begin{lemma}\label{lemma:lift}
    Define the map $\Lambda: \calM \times (\R^3\times\R^3)\to\rslam$ by
    \begin{align}\label{eq:lift}
        \Lambda(&(P,  q_i), (\omega, a)):= \\
        &\left((U+D)+P^{-1}(G-D)P,\left(\omega+\frac{q_i^\times R_P^{\top} v_P}{q_i^{\top} q_i}, \frac{q_i^{\top} R_P^{\top} v_P}{q_i^{\top} q_i}\right)\right),\notag
    \end{align}
    where
    \begin{align*}
        U &= \begin{pmatrix}
            \omega^\times & a & \mathbf{0}_{3 \times 1}\\
            \mathbf{0}_{1 \times 3} & 0 & 0 \\
            \mathbf{0}_{1 \times 3} & 0 & 0
        \end{pmatrix}, &
        D &=\left(\begin{array}{ccc}
            \mathbf{0}_{3 \times 3} & \mathbf{0}_{3 \times 1} & \mathbf{0}_{3 \times 1} \\
            \mathbf{0}_{1 \times 3} & 0 & 1 \\
            \mathbf{0}_{1 \times 3} & 0& 0
            \end{array}\right), \\
        G &= \begin{pmatrix}
            \mathbf{0}_{3 \times 3} & g\mathbf{e}_3 & \mathbf{0}_{3 \times 1}\\
            \mathbf{0}_{1 \times 3} & 0 & 0 \\
            \mathbf{0}_{1 \times 3} & 0 & 0
        \end{pmatrix}.
    \end{align*}
    Then $\Lambda$ is a lift \cite{van2022equivariant} of the system dynamics~\eqref{eq:system_dynamics}; that is, it satisfies the lift condition
    \begin{align}\label{eq:lift_condition}
        \left.\mathrm{D}_E\right|_{\mathrm{id}} \phi_{\left(P, q_i\right)}(E) \left[\Lambda\left(\left(P, q_i\right),(\omega, a)\right)\right]=f_{(\omega, a)}\left(P, q_i\right).
    \end{align}
\end{lemma}

\vspace*{2mm}
\subsection{Origin choice and local coordinates}
The equivariant filter requires a choice of the origin on the state space and a set of local coordinates.
In this paper, we use the \emph{log-polar} parametrisation proposed in \cite{van2019geometric} that is associated with normal coordinates on the Lie group, noting that this is closely aligned with the polar coordinates investigated in \cite{djugash2009robust}.
Let $\mr{\xi} = (\mr{P},\mr{q}_i)\in\calM$ denote the fixed choice of origin.
We leave the particular choice of the pose origin $\mr{P} \in \calSE_2(3)$ arbitrary but fix $\mr{q}_i:=\mathbf{e}_3$ for all $i$.
Define the normal coordinates $\vartheta:U_{\mr{\xi}}\subset\calM\to\R^{9+3n}$ as
\begin{align}\label{eq:normal_coordinate}
    \vartheta(P, q_i) = \begin{pmatrix}
        \log_{\SE_2(3)}(\mr{P}^{-1}P)\\
        \varsigma_{\SOT(3)}(q_1)\\
        \vdots\\
        \varsigma_{\SOT(3)}(q_n)
    \end{pmatrix},
\end{align}
where the $\SOT(3)$ normal coordinates $\varsigma_{\SOT(3)}:\R^3\to\R^3$ are defined as \cite{van2023eqvio}
\begin{align}
   &\varsigma_{\SOT(3)}(q):=\left(\begin{array}{c}
        \arccos \left(\frac{q_3}{|q|}\right) \frac{q_2}{\left|\mathbf{e}_3 \times q\right|} \\
        \arccos \left(\frac{q_3}{|q|}\right) \frac{-q_1}{\left|\mathbf{e}_3 \times q\right|} \\
        -\log (|q|)
        \end{array}\right)
\end{align}
Here, $\vartheta$ serves as the coordinate chart for $\calM$ in $U_{\mr{\xi}}$, a neighbourhood of $\mr{\xi}$.
It provides normal coordinates for the state space $\calM$ with respect to the state action $\phi$.

\subsection{EqF Dynamics}
Following the EqF design \cite{van2022equivariant,ge2022equivariant}, let $\xi = (P, q_i)\in\calM$ denote the true state of the system, and $\hat{X} = (\hat{T}, \hat{Q}_i)\in\RSLAM$ denote the observer state.
The state estimate is defined by applying the group action $\hat{\xi} = \phi(\hat{X}, \mr{\xi})$.
The equivariant error is $e:=\phi(\hat{X}^{-1}, \xi)$, and the local state error is defined by $\varepsilon:=\vartheta(e)$ as long as $e \in \calU_{\mr{\xi}}$.

Let $(\omega_m, a_m)\in\R^3\times\R^3$ denote the measured IMU data and $y\in\R_+^n$ denote the measured range data. Let $\Sigma\in\mathbb{S}_+(9+3n)$ be the Riccati matrix where $\mathbb{S}_+(9+3n)$ denotes the set of positive definite $(9+3n)\times(9+3n)$ matrices.
Define $A_t$, $B_t$ and $C_t^\star$ to be the Jacobian matrices of the EqF, defined in Appendix~\ref{sec:eqf_matrices}.
Then the filter dynamics are given by

\small
\begin{align}\label{eq:observer_dynamics}
    &\dot{\hat{X}} = \hat{X}\Lambda(\phi(\hat{X},\mr{\xi}), (\omega_m, a_m)) - \Delta\hat{X}, \qquad \hat{X}(0) = \id,\\
    & \Delta=\tD_E\big|_{\mathrm{id}} \phi_{\mr{\xi}}(E)^{\dagger} \cdot \tD_{\xi}\big|_{\mr{\xi}} \vartheta(\xi)^{-1} \Sigma {C_t^\star}^\top N_t^{-1}(y-h(\hat{\xi})), \\
    & \dot{\Sigma}=A_t \Sigma+\Sigma A_t^{\top}+B_t M_t B_t^{\top}-\Sigma {C_t^\star}^\top N_t^{-1} C_t^\star \Sigma -\Gamma_{\mathrm{D} \phi_{\mr{\xi}} \Delta} \Sigma-\Sigma \Gamma_{\mathrm{D} \phi_{\mr{\xi}} \Delta}^{\top},
\end{align}\normalsize
where:
\begin{itemize}
    \item $\Delta\in\rslam$ is the correction term;
    \item $\tD_E\big|_{\mathrm{id}} \phi_{\mr{\xi}}(E)^{\dagger}$ is a right-inverse of the differential of the state action at the origin;
    \item $M_t\in\mathbb{S}_+(6)$ and $N_t\in\mathbb{S}_+(n)$ are the input and output gain matrices;
    \item $\Gamma_{\mathrm{D} \phi_{\mr{\xi}} \Delta}$ is the modification term for the reset step \cite{ge2022equivariant}.
\end{itemize}

\section{EXPERIMENTS}\label{sec:experiment}
We test the proposed EqF on a real-world dataset collected using a fleet of aerial and ground vehicles \cite[Chapter 7]{Henderson_2024}.
The vehicles are traversing long paths in an outdoor field with a number of static landmarks.
We present the performance of the EqF on a full SLAM task, where the landmarks are initialised with no bearing information.
There is no pre-filtering or initialisation phase implemented. 
The goal of this experiment is for the vehicle to localise itself and estimate the relative positions of the landmarks using only the IMU and range sensor data.

\begin{figure}[htb!]
    \centering
    \includegraphics[width=\linewidth]{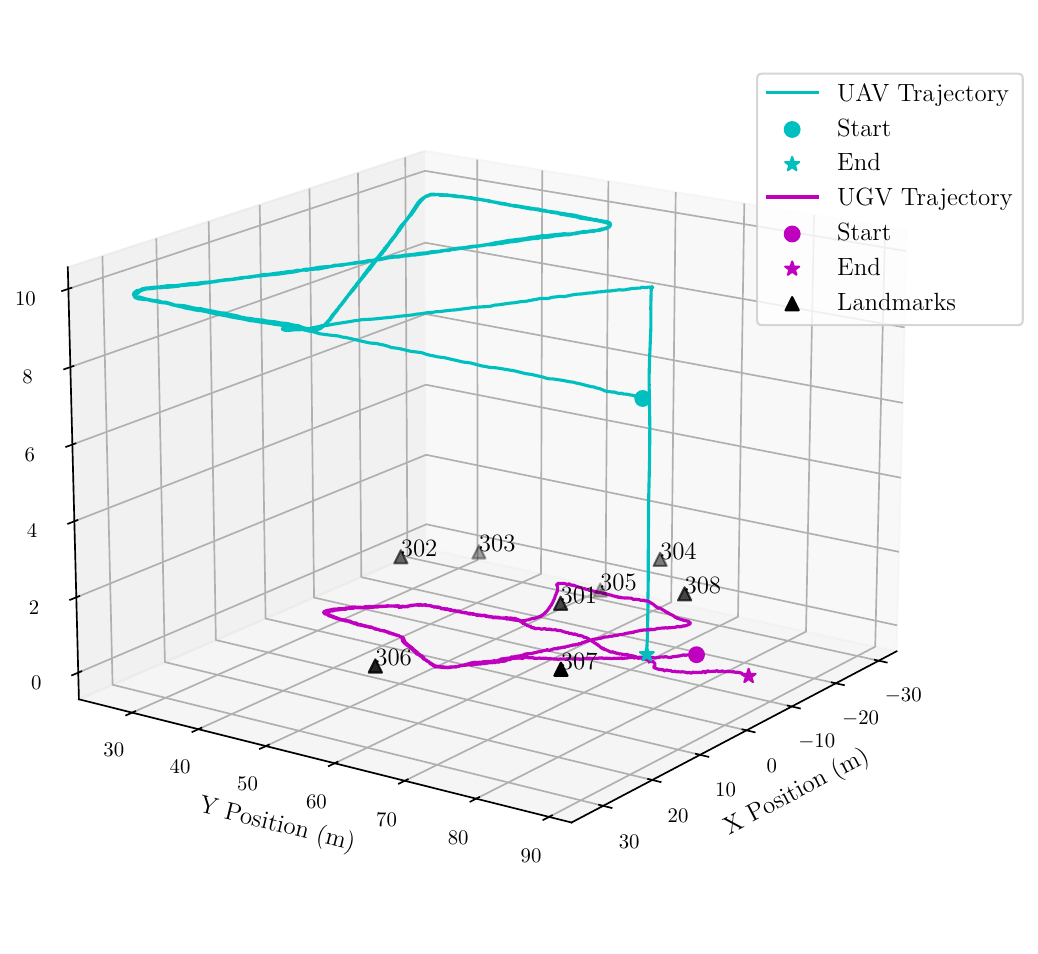}
    \vspace*{-1.2cm}
    \caption{Visualisation of the selected vehicle trajectories (Mission 1 UAV 1, Mission 2 UGV 1) and the landmark positions. }
    \label{fig:test_trajectory}
\end{figure}

\subsection{Hardware Setup}
For a complete description of the dataset and sensor characterisation, we refer the reader to \cite[Chapter 7]{Henderson_2024}.

The dataset is recorded on a fleet of three aerial and three ground vehicles, traversing multiple paths in an outdoor field with eight static landmarks.
Each vehicle is equipped with an RTK-GPS receiver which provides highly accurate (2cm) positioning data.
Two onboard IMUs (ICM-20689 and BMI088) provide angular velocity and acceleration data at 400Hz.
The range measurement data is collected using a UWB sensor (Decawave DWM1001-DEV) at 10Hz, providing distance measurements between the vehicle and the landmarks.
No inter-landmark information is used.
For the data considered, the operational distance was around 50m and the observed standard deviation of the range measurement error is around 0.25m.
Due to the number of independent vehicles operating radio links using the same frequency band as the UWB sensors, the experiment suffered from significant dropout in the range data (commonly 10s gaps in data).
To compensate, we have synthesised data from the ground truth (RTK-GPS) and used other independent experiments from the same sensors to provide real noise characteristics on the synthesised signals.
We run the algorithms on four different sequences, including two aerial and two ground vehicle missions, to demonstrate the performance of the proposed EqF under different conditions.

\subsection{Filter Implementation}
For comparison, we implement a standard EKF which uses Euclidean coordinates to model the landmarks, as presented in \cite{kantor2002preliminary}.
The original work models the navigation states in Euclidean space which is not directly comparable to the EqF.
To make the comparison fair, we use the same $\SE_2(3)$ structure for the navigation states to remove the effect of different linearisation schemes in the filter propagation, focusing on the symmetry of the range measurements.
Both EqF and EKF are initialised with the same initial Riccati matrix and gain matrices for the navigation states.
We do not correct the IMU biases for the dataset, and neither filter has bias states in the system model.

In both filters, there is no prior information for the landmarks available.
Each landmark is initialised in the vehicle body-fixed frame using the first range measurement with a fixed direction $\mathbf{e}_3$.
There is no additional pre-filter process to initialise the landmarks.
For the EqF, the initial covariance of the landmarks is set to be $\text{diag}(3.0, 3.0, 3.0)$ in units $(\text{rad}^2, \text{rad}^2, \log(\text{m})^2)$, for both the aerial and ground vehicles.
This corresponds to roughly 45 degree uncertainty in the bearing of the landmarks, and 0.2m in range.  
We use an initial covariance of $\text{diag}(50.0, 50.0, 50.0)$ in units $(\text{m}^2, \text{m}^2, \text{m}^2)$ for the EKF on aerial vehicles.
This corresponds to an uncertainty of roughly 8m in each axis of Euclidean coordinates. 
Noting that the vehicle flight path is 10m high and the UWB sensors are on poles 2m high, a 45 degree uncertainty in landmark bearing is roughly an 8m uncertainty in the horizontal coordinates of the landmark.
For the ground vehicles, we use an initial covariance of $\text{diag}(10.0, 10.0, 10.0)$ in units $(\text{m}^2, \text{m}^2, \text{m}^2)$.
Since the EKF state representation does not coincide with the geometry of the range sensor, a much larger initial covariance in range required for the EKF to compensate for linearisation error in the filter during transients.
Since there is no prior information on landmark position the raw result from the EqF is not globally aligned.
In order to compare the results with the groundtruth, we apply a simple $\SE(3)$ affine transform to the final solution of the path and map estimates to align the results using Umeyama method \cite{umeyama1991least}.

\subsection{Results}
Table \ref{tab:rmse_comparison} and \ref{tab:mapping_error} show the RMSE of the robot path and landmark mapping error for the EKF and EqF, on four different sequences.
For the path error, we report the translational RMSE of both the last 40\% of the trajectory and the whole trajectory.
As shown in the tables, the EqF clearly outperforms the EKF in all cases.
The EKF fails to converge on one of the sequences (highlighted in \textcolor{red}{red}), giving large error in the landmark estimation.

\begin{table}[h]
    \centering
    \subfloat[Whole Trajectory]{
        \begin{tabular}{lcccc}
            \toprule
             & M1 UAV1 & M3 UAV1 & M1 UGV1 & M2 UGV1 \\
            \midrule
            EKF & 6.710 & {10.892} & \textcolor{red}{10.653} & 6.407 \\
            EqF & \textbf{0.655} & \textbf{1.543} & \textbf{2.563} & \textbf{1.522} \\
            \bottomrule
        \end{tabular}
    }\hfill

    \subfloat[Last 40\% of the Trajectory]{
        \begin{tabular}{lcccc}
            \toprule
            & M1 UAV1 & M3 UAV1 & M1 UGV1 & M2 UGV1 \\
            \midrule
            EKF & 2.889 & {5.233} & \textcolor{red}{9.479} & 0.618 \\
            EqF & \textbf{0.323} & \textbf{0.235} & \textbf{0.304} & \textbf{0.597} \\
            \bottomrule
        \end{tabular}
    }
    \caption{Translational RMSE in unit (m) Comparison for EKF and EqF. \textbf{Bold} indicates the best result. \textcolor{red}{Red} indicates that the filter fails to converge.}
    \label{tab:rmse_comparison}
\end{table}

\begin{table}[h]
    \centering
    
    \begin{tabular}{lcccc}
        \toprule
        & M1 UAV1 & M3 UAV1 & M1 UGV1 & M2 UGV1 \\
        \midrule
        EKF & 1.170 (0.17) & {2.870 (0.26)} & \textcolor{red}{8.377 (0.63)} & 1.484 (0.43) \\
        EqF & \textbf{0.185 (0.11)} & \textbf{0.153 (0.08)} & \textbf{0.581 (0.22)} & \textbf{0.379 (0.20)} \\
        \bottomrule
    \end{tabular}
    \caption{Mapping Error Comparison for EKF and EqF. The standard deviation is listed next to the mean, both in unit (m). \textbf{Bold} indicates the best result. \textcolor{red}{Red} indicates that the filter fails to converge.}
    \label{tab:mapping_error}
\end{table}

\begin{figure}[htb!]
    \centering
    \subfloat[Mission 1 UAV 1]{%
        \includegraphics[width=0.48\linewidth]{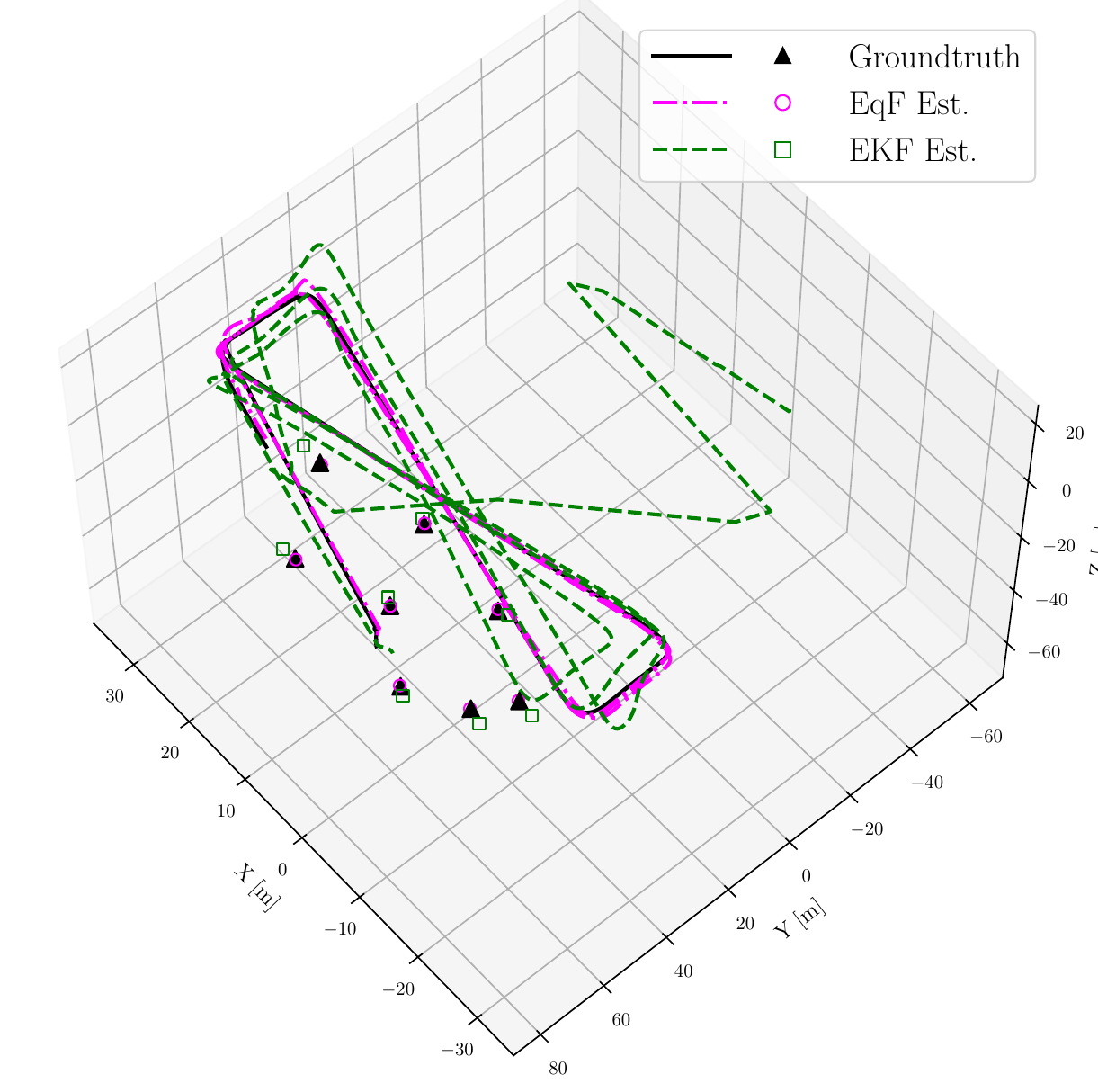}
        \label{fig:slam_ekf}}
    \hfill
    \subfloat[Mission 2 UGV 1]{%
        \includegraphics[width=0.48\linewidth]{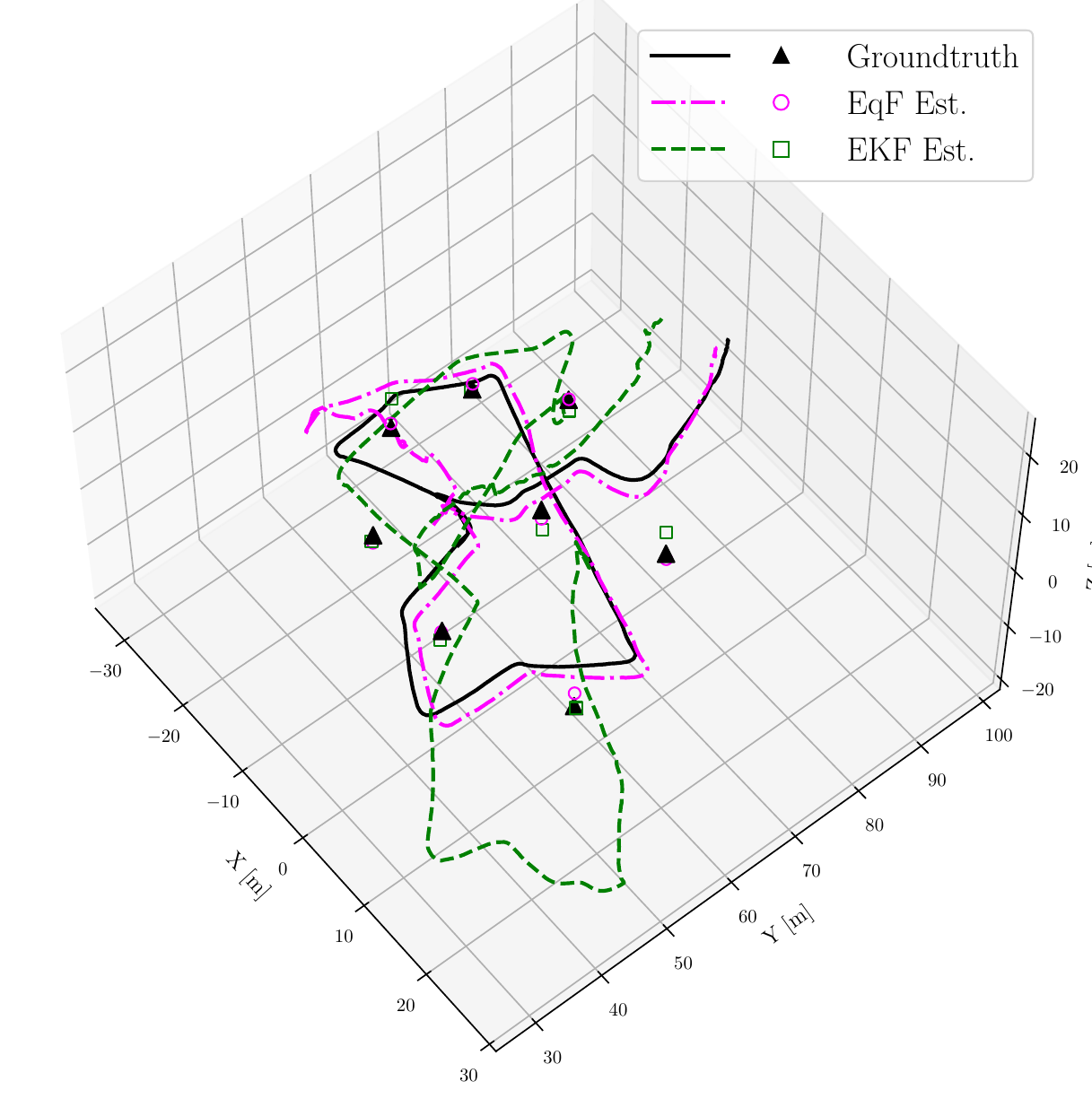}
        \label{fig:slam_eqf}}
    \caption{The true and estimated trajectories of the robot and the landmark positions after alignment. The axes are adjusted for better visualisation.}
    \label{fig:slam_trajectory}
\end{figure}

\begin{figure}[htb!]
    \centering
    \subfloat[Landmark position error using EKF.]{%
        \includegraphics[width=\linewidth]{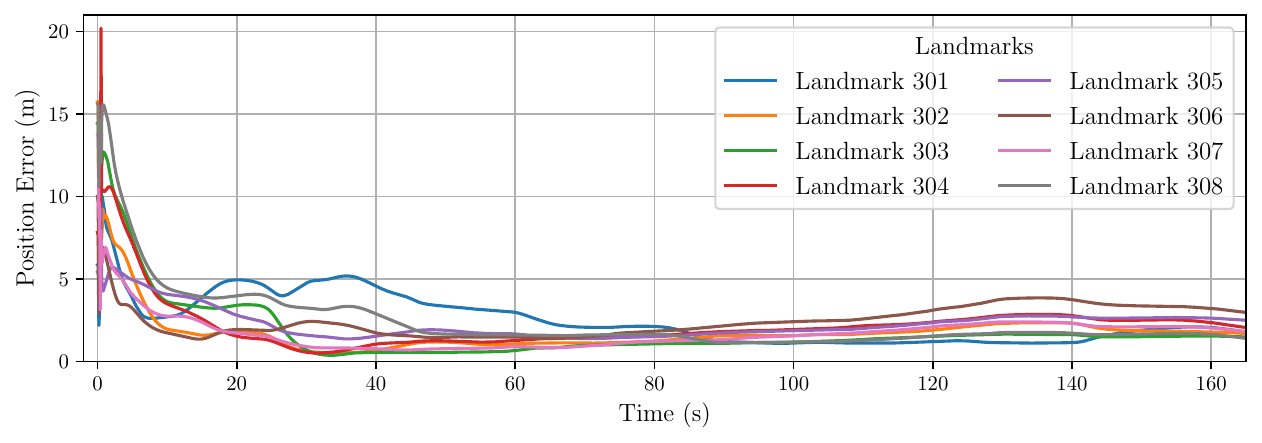}
        \label{fig:slam_error_euc}}
    \hfill
    \subfloat[Landmark position error using EqF.]{%
        \includegraphics[width=\linewidth]{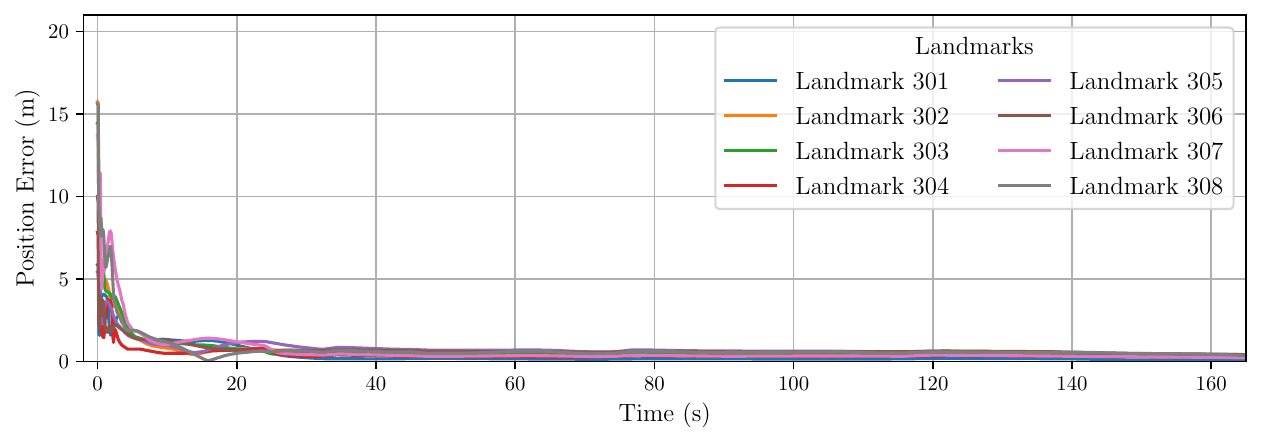}
        \label{fig:slam_error_normal}}
    \caption{Estimation error of landmark positions on Mission 1 UAV 1.}
    \label{fig:slam_error}
\end{figure}

To further analyse the performance of EqF against the EKF, we pick two of the sequences, collected from an aerial vehicle (Mission 1 UAV 1) and a ground vehicle (Mission 2 UGV 1) respectively, and visualise the estimated robot path and landmark positions.
Their trajectories are shown in Figure~\ref{fig:test_trajectory}.
In Figure~\ref{fig:slam_trajectory}, we present the output of the proposed EqF and the EKF for the robot path and landmark positions after applying the affine transform that best aligns the results with the groundtruth.
The landmark error convergence on Mission 1 UAV 1 is shown in Figure~\ref{fig:slam_error}.

As shown in Figure \ref{fig:slam_eqf}, the proposed EqF is able to converge quickly in a no prior situation for the SLAM task, and clearly outperforms the EKF in both the RMSE of the robot path and the mapping error of the landmarks.
In terms of the landmark error convergence (see Figure~\ref{fig:slam_error}), the EqF managed to localise the landmarks to within 1-2m of the true positions in the first 10 seconds, while the EKF cannot reach the same level of accuracy even by the end of the data.
This also significantly improves the EqF's robot path estimation, both of the full path and the final 40\%.
Additionally, because of the covariance being modelled in the $\SOT(3)$ normal coordinates, the EqF is able to converge with a much smaller initial covariance matrix which is independent of the distance to the landmarks, while the EKF requires a big initial covariance to search for the landmarks.
This leads to the different behaviors in the initial transient phase of the filters, where the EqF converges in a much faster and smoother manner, as shown in Figure~\ref{fig:slam_trajectory}.

\section{CONCLUSION}
This paper makes the following contributions:
\begin{itemize}
\item We propose a novel method for RO-SLAM that uses the symmetry group $\RSLAM$, which we prove to be compatible with the system dynamics and the measurement function.
\item We use the $\SOT(3)$ normal coordinates to model the landmark positions in the body-fixed frame, which, similar to prior polar coordinate state representations, aligns closely with the nonlinear distribution of landmarks information states that results from range measurements.
\item We design an equivariant filter for RO-SLAM using the proposed symmetry group.
    The performance of the filter is demonstrated on a real-world dataset, and is shown to be able to localise the vehicle and estimate the landmark positions accurately without prior information or bootstrapping.
\end{itemize}

\appendix
\subsection{Symmetry group proofs}\label{sec:proofs}
\paragraph{Proof of Lemma~\ref{lemma:output_action}}
    Consider the output equivariance condition~\eqref{eq:output_equivariance_condition}, choose some arbitrary $(T, Q_i)\in\RSLAM$ and $(P,  q_i)\in\calM$, one has
    \begin{align*}
        h(\phi((T, Q_i), (P,q_i)))  &= h((PT,  c_{Q_i}^{-1}R_{Q_i}^\top q_i))\\
                                    &= (\lVert  c_{Q_i}^{-1}R_{Q_i}^\top q_i\rVert)
                                    =(c_{Q_i}^{-1} \lVert q_i \rVert)\\
                                    &=\rho((T, Q_i),(\lVert q_i \rVert))\\
                                    &=\rho((T, Q_i), h(P,  q_i)).
    \end{align*}
    This completes the proof.

\paragraph{Proof of Lemma~\ref{lemma:lift}}
    Consider the lift condition~\eqref{eq:lift_condition}, choose an arbitrary state $(P, q_i)\in\calM$ and input $(\omega, a)\in(\R^3\times\R^3)$, one has
    \begin{align*}
        &\quad\left.\mathrm{D}_E\right|_{\mathrm{id}} \phi_{\left(P, q_i\right)}(E) \left[\Lambda\left(\left(P, q_i\right),(\omega, a)\right)\right]\\
        &= \tD_{(T,Q_i)}\big|_{(I_5, I_4)} (PT, c^{-1}_{Q_i}R_{Q_i}^\top q_i)\\
        &\quad\left[\left((U+D)+P^{-1}(G-D)P,\left(\omega+\frac{q_i^\times R_P^{\top} v_P}{q_i^{\top} q_i}, \frac{q_i^{\top} R_P^{\top} v_P}{q_i^{\top} q_i}\right)\right)\right].
    \end{align*}
    The proof of the lift condition related to the navigation state can be found in \cite{fornasier2023equivariant}.
    In this paper, we only prove the lift condition for the part associated with the landmarks.

    Consider the part of the group action $\phi$ where $\SOT(3)\subset\RSLAM$ element acts on $\R^3\subset\calM$, one has
    \begin{align*}
        &\;\tD_{Q_i}\big|_{I_4}(c^{-1}_{Q_i}R_{Q_i}^\top q_i)\left[\left(\omega+\frac{q_i^\times R_P^{\top} v_P}{q_i^{\top} q_i}, \frac{q_i^{\top} R_P^{\top} v_P}{q_i^{\top} q_i}\right)\right]\\
        &=\left.\ddt\right|_{t=0} \left(1+\frac{q_i^{\top} R_P^{\top} v_P}{q_i^{\top} q_i}\right)^{-1}\left(I_3+t\left(\omega+\frac{q_i^\times R_P^{\top} v_P}{q_i^{\top} q_i}\right)^\times\right)q_i\\
        &= -\frac{q_i^{\top} R_P^{\top} v_P}{q_i^{\top} q_i}q_i - \left(\omega+\frac{q_i^\times R_P^{\top} v_P}{q_i^{\top} q_i}\right)^\times q_i\\
        &=-\omega^\times q_i - -\frac{q_i q_i^{\top} R_P^{\top} v_P}{q_i^{\top} q_i} + \frac{q_i^\times q_i^\times R_P^{\top} v_P}{q_i^{\top} q_i}\\
        &=-\omega^\times q_i - \frac{q_i q_i^\top - q_i^\times q_i^\times }{q_i^{\top} q_i} R^\top_P v_P = -\omega^\times q_i - R^\top_P v_P,
    \end{align*}
    where the last equality follows from the property
    \[ {\left(\frac{q_i}{\lVert q_i\rVert}\right)^\times}^2 = \frac{q_i}{\lVert q_i\rVert}\frac{q_i}{\lVert q_i\rVert}^\top - I_3 = \frac{q_i q_i^\top}{q_i^\top q_i} - I_3.\]
    Recall the dynamics of the landmark position $q_i$ given in \eqref{eq:q_dyanmics}, this completes the proof.

\subsection{EqF Matrices}\label{sec:eqf_matrices}
In this section, we present the Jacobian matrices used in the filter implementation.
Let $\mr{\xi}=(\mr{P}, \mr{q}_i)\in\calM$ denote the choice of the origin.
We leave the term $\mr{P}\in\SE_2(3)$ arbitrary, and fix $\mr{q}_i:=\mathbf{e}_3$ for all $i$.
Let $\hat{X}=(\hat{T}, \hat{Q}_i)\in\RSLAM$ denote the observer state, and $\hat{\xi}=(\hat{P}, \hat{q}_i)\in\calM$ denote the current state estimate given by \eqref{eq:state_action} with $\hat{P} = (\hat{R}, \hat{x}, \hat{v})$.
The input is denoted by $u=(\omega, a)\in\R^3\times\R^3$.
The state matrix $A_t$ is given by\small
\begin{align*}
    A_t = &\tD_e \big|_{\mr{\xi}} \vartheta(e)\cdot \tD_\xi \big|_{\hat{\xi}} \phi_{\hat{X}^{-1}}(\xi) \cdot \tD_E \big|_{\text{id}} \phi_{\hat{\xi}}(E)\cdot \tD_\xi \big|_{\hat{\xi}} \Lambda(\xi, u)\\
    & \qquad\qquad\qquad\qquad\qquad\qquad\cdot \tD_e \big|_{\mr{\xi}} \phi_{\hat{X}}(e) \cdot \tD_\varepsilon \big|_{0} \vartheta^{-1}(\varepsilon)\\
    =& \begin{pNiceArray}{ccc:ccc}[margin]
        \mathbf{0} &\mathbf{0} &\mathbf{0} &\Block{3-3}{\mathbf{0}_{9 \times 3n}}&&  \\
        \mathbf{0}& \mathbf{0}& I_3& & & \\
        g^\times&\mathbf{0} &\mathbf{0} &&& \Bstrut\\
        \hdottedline
        \Block{1-2}{\mathbf{0}_{3 \times 6}} & & A_{q_1 v} & A_{q_1 q_1} & \hdots & \mathbf{0}_{3\times 3} \Tstrut\\
        \Block{1-2}{\vdots} && \Block{1-1}{\vdots} & \vdots & \ddots & \vdots \\
        \Block{1-2}{\mathbf{0}_{3 \times 6}} & & A_{q_n v} &\mathbf{0}_{3\times 3} &\hdots & A_{q_n q_n} \\
    \end{pNiceArray}\in \R^{(9+3n) \times (9+3n)},
\end{align*}\normalsize
with \small
\begin{align*}
A_{q_i v} &:= \tD_e \big|_{\mr{\xi}} \vartheta(e)\cdot\hat{Q}_i\left(\hat{q}^\times_i\frac{\hat{q}_i^\times\hat{R}^\top}{\hat{q}_i^\top\hat{q}_i} + \hat{q}_i\frac{\hat{q}_i^\top\hat{R}^\top}{\hat{q}_i^\top\hat{q}_i}\right),\\
A_{q_i q_i} &:= -\tD_e \big|_{\mr{\xi}} \vartheta(e)\cdot\hat{Q}_i\left(\hat{q_i}^\times\left(\frac{(\hat{R}^\top\hat{v})^\times}{\hat{q}_i^\top\hat{q}_i} + \frac{2\hat{q}_i^\times\hat{R}^\top\hat{v}\hat{q}_i^\top}{(\hat{q}_i^\top\hat{q}_i)^2}\right)\right.\\
&\qquad+\hat{q}_i\left( \frac{(\hat{R}^\top\hat{v})^\top}{\hat{q}_i^\top\hat{q}_i} - \frac{2\hat{q}_i^\top\hat{R}^\top\hat{v}\hat{q}_i^\top}{(\hat{q}_i^\top\hat{q}_i)^2}\right)\hat{Q}_i^{-1}\cdot \tD_\varepsilon \big|_{0} \vartheta^{-1}(\varepsilon).
\end{align*}\normalsize

The input matrix $B_t$ is given by \small
\begin{align*}
    B_t = &\tD_e \big|_{\mr{\xi}} \vartheta(e)\cdot \tD_\xi \big|_{\hat{\xi}} \phi_{\hat{X}^{-1}}(\xi) \cdot \tD_E \big|_{\text{id}} \phi_{\hat{\xi}}(E)\cdot \tD_u \big|_{(\omega,a)} \Lambda(\hat{\xi}, u)\\
    =& \begin{pNiceArray}{cp{1.2cm}:c}[margin]
        \Block{1-2}{\hat{R}} & & \mathbf{0}\\
        \Block{1-2}{\hat{p}^\times\hat{R}} & & \hat{R} \\
        \Block{1-2}{\hat{v}^\times\hat{R}} & & \mathbf{0} \\
        \hdottedline
        \Block{1-2}{B_{q_1\omega}} && \Block{1-1}{\mathbf{0}}\Tstrut\\
        \Block{1-2}{\vdots} & & \Block{1-1}{\vdots} \\
        \Block{1-2}{B_{q_n\omega}} & & \Block{1-1}{\mathbf{0}}
    \end{pNiceArray}\in \R^{(9+3n) \times 6},
\end{align*}\normalsize
with $B_{q_i \omega}:=\tD_e \big|_{\mr{\xi}} \vartheta(e)\cdot\hat{Q}_i \hat{q_i}^\times\in\R^{3\times 3}$.
The differential of coordinate transformation $\tD_e \big|_{\mr{\xi}} \vartheta(e)$ is given by $\tD_e \big|_{\mr{\xi}} \vartheta(e) = \begin{pmatrix}
    0 & 1 & 0\\
    -1 & 0 & 0\\
    0 & 0 & -1
\end{pmatrix}$.

The equivariant output matrix $C_t^\star$ is given by \small
\begin{align*}
    C_t^{\star} \varepsilon&=\frac{1}{2}\left(\mathrm{D}_{E \mid \mathrm{id}} \rho(E, y)+\mathrm{D}_{E \mid \mathrm{id}} \rho(E, \hat{y})\right) \operatorname{Ad}_{\hat{X}^{-1}} \varepsilon^{\wedge}\\
    C_t^\star &= \begin{pNiceArray}{cp{0.8cm}c:cccc}[margin]
        \Block{4-3}{\mathbf{0}_{n \times 9}}& &  &C^\star_{q_1} & 0 & \hdots & 0 \\
        & & & 0 & \ddots & & \vdots \\
        & & & \vdots & & \ddots & 0 \\
        & & & 0 & \hdots & 0 & C^\star_{q_n}
        \end{pNiceArray}\in \R^{n \times (9+3n)},
\end{align*}\normalsize
where each $C^\star_{q_i}\in\R^{1\times 3}$ is given by $C_{q_i}:= \left(0\quad 0 \quad -\frac{1}{2}(y_{q_i}+\lvert \hat{q}_i\rvert)\right)$.

\bibliography{reference}
\bibliographystyle{IEEEtran}

\end{document}